\pdfoutput=1

\documentclass[11pt]{article}

\usepackage[final]{acl}

\usepackage{times}
\usepackage{latexsym}
\usepackage{microtype}
\usepackage{hyperref}
\usepackage{url}
\usepackage{booktabs}
\usepackage{subcaption}
\definecolor{darkblue}{rgb}{0, 0, 0.5}
\hypersetup{colorlinks=true, citecolor=darkblue, linkcolor=darkblue, urlcolor=darkblue}

\usepackage{algorithm}
\usepackage{makecell}
\usepackage{CJKutf8}
\usepackage{color}
\usepackage{amssymb}
\usepackage{amsmath}
\usepackage{bbding}
\usepackage{algpseudocode}
\usepackage{algorithmicx,algorithm}

\usepackage{multirow}
\usepackage{booktabs}
\usepackage{arydshln}
\usepackage{graphicx}
\newcommand{\chen}[1]{{\color{black} #1}}
\usepackage[T1]{fontenc}

\usepackage[utf8]{inputenc}

\usepackage{microtype}

\usepackage{inconsolata}

\usepackage{graphicx}

\title{SGIC: A Self-Guided Iterative Calibration Framework for RAG}

\author{
 \textbf{Guanhua Chen$^1$\thanks{Equal contribution}},
 \textbf{Yutong Yao$^1$\footnotemark[1]},
 \textbf{Lidia S. Chao$^1$},
 \textbf{Xuebo Liu$^2$},
 \textbf{Derek F. Wong$^1$\thanks{Corresponding authors.}}
\\
 $^1$NLP\textsuperscript{2}CT Lab, Department of Computer and Information Science, University of Macau
\\
$^2$Institute of Computing and Intelligence, Harbin Institute of Technology, Shenzhen, China
\\
\{nlp2ct.guanhua, nlp2ct.yutong\}@gmail.com\\
liuxuebo@hit.edu.cn\\
\{derekfw, lidiasc\}@um.edu.mo\\
}

\begin{document}
\maketitle
\begin{abstract}
Recent research in retrieval-augmented generation (RAG) has concentrated on retrieving useful information from candidate documents. However, numerous methodologies frequently neglect the calibration capabilities of large language models (LLMs), which capitalize on their robust in-context reasoning prowess. This work illustrates that providing LLMs with specific cues substantially improves their calibration efficacy, especially in multi-round calibrations. We present a new \textbf{SGIC}: \textbf{S}elf-\textbf{G}uided \textbf{I}terative \textbf{C}alibration Framework that employs uncertainty scores as a tool. Initially, this framework calculates uncertainty scores to determine both the relevance of each document to the query and the confidence level in the responses produced by the LLMs. Subsequently, it reevaluates these scores iteratively, amalgamating them with prior responses to refine calibration. Furthermore, we introduce an innovative approach for constructing an iterative self-calibration training set, which optimizes LLMs to efficiently harness uncertainty scores for capturing critical information and enhancing response accuracy. Our proposed framework significantly improves performance on both closed-source and open-weight LLMs.
\end{abstract}

\section{Introduction}
Retrieval-augmented generation (RAG) necessitates intricate reasoning across various candidate-retrieved documents with more than one hop. The advent of exceptionally large language models (LLMs) such as GPT-3.5 and Claude has markedly augmented the capabilities of robust in-context reasoning prowess, enabling significant advances in RAG performance via in-context learning or multi-step reasoning \cite{wang-etal-2023-self-prompted, khalifa2023few}, without additional training. Despite these enhancements, the deployment of such LLMs in local settings is frequently impractical, because of their proprietary nature and voluminous parameter sets. Consequently, researchers focus on fine-tuning open-weight LLMs to improve performance in specific downstream applications \cite{zheng2023judging, du2022glm, DBLP:conf/nips/Liu0W0WH024}.

\begin{figure*}[t]
    \centering
    \begin{subfigure}[b]{0.45\textwidth}
     \includegraphics[width=1\linewidth]{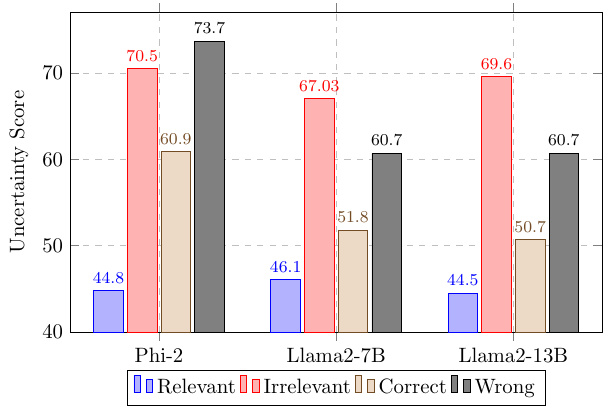}
    \caption{Three different baseline models. }
    \label{fig:sub1}
  \end{subfigure}
  \hfill 
  \begin{subfigure}[b]{0.45\textwidth}
     \includegraphics[width=1\linewidth]{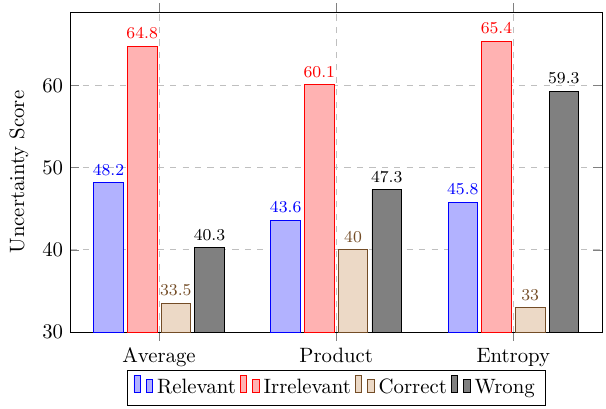}
    \caption{Three different uncertainty estimations.}
    \label{fig:sub2}
  \end{subfigure}
    \caption{The uncertainty score of the relevant/irrelevant documents and correct/wrong generated answers on 2,000 samples extracted from HotpotQA\cite{yang-etal-2018-hotpotqa} dataset.}
    \label{fig:1}
\end{figure*}

However, lots of existing works on RAG primarily concentrate on extracting relevant documents or the refinement of specialized instructions \cite{asai2022task, ziems2023large, wang2023query2doc, sun2023chatgpt, ma2023zero, tang2023found}, which does not fully leverage the in-context reasoning abilities intrinsic to LLMs. Inspired by recent studies that employ LLMs for self-calibration to generate better answers through self-feedback mechanisms \cite{peng2023check, dhuliawala2023chain, shinn2023reflexion}, we argue that LLMs can optimize generated answers of RAG by utilizing prior responses coupled with strategic hints to facilitate in-context reasoning. The preliminary experimental findings, detailed in Section \ref{sec.tag}, lend credence to our hypothesis. In addition, Figure \ref{fig:sub1} reveals a distinct gap in the uncertainty scores produced by LLMs when distinguishing correct/incorrect answers and the relevant/irrelevant documents. Meanwhile, several widely used uncertainty estimation approaches (Figure \ref{fig:sub2}) confirm the generality of this phenomenon. 

Thus, we introduce a novel \textbf{SGIC}: \textbf{S}elf-\textbf{G}uided \textbf{I}terative \textbf{C}alibration Framework that harnesses the robust in-context reasoning capabilities of LLMs to self-calibrate previously generated answers. Inspired by \citet{DBLP:conf/emnlp/0004L0ZWC024}, we adopt the uncertainty scores in the inference stage to iteratively rephrase the input prompts, incorporating the uncertainty scores of previously generated answers to steer the LLMs towards in-context reasoning and self-calibration. Additionally, we introduce document uncertainty scores to assess the relevance between each document and the question, assisting LLMs in retrieving the most pertinent documents. For small-scale models with limited long-document understanding capabilities, we further propose a strategy for reconstructing a self-calibration training dataset following the procedure in Figure \ref{fig:2}, to fine-tune the LLMs to utilize the uncertainty scores of the document and previous responses to generate a better answer.

Empirically, we evaluate our framework across two benchmarks, HotpotQA \cite{yang-etal-2018-hotpotqa} and Natural Question (NQ) \cite{kwiatkowski-etal-2019-natural}, on two strong close-source LLMs, \textbf{GPT-4o} and \textbf{GPT-4o mini} \cite{achiam2023gpt} and two open-weight LLMs, which are \textbf{Phi-3.5} \cite{textbooks2} and \textbf{Llama2-7B-Chat} \cite{touvron2023llama}. The experimental results demonstrate the validity and strong potential of our framework.

\section{Related Work}
\subsection{Retrieval-Augmented Generation (RAG)}

RAG retrieves relevant documents from a knowledge base and employs a generator to produce coherent and accurate responses based on the retrieved documents \cite{lewis2020retrieval}. Recent studies \cite{jiang2023active, chen2024benchmarking, fan2024survey} have also demonstrated that RAG can effectively address the hallucination and incorrect reasoning problems of LLMS in the Question Answering (QA) and downstream tasks, such as Document Question Answering (DQA), which require LLMS to reason between multiple documents. \citet{trivedi2022interleaving} mutually integrated the Chain-of-Thoughts (CoT) into the retrieval step to enhance the retrieval capability of LLMs for multi-hop QA. 
\citet{ma2023query} proposed query rewriting RAG through Adaptive queries generated by a fine-tuned rewriter.
\citet{ma2023large} proposed an adaptive filter-then-rerank paradigm, prompting LLMs to rerank few-shot hard samples filtered by small LMs, which enhances the perception of key information of the LLMs.
\citet{jeong2024adaptive} proposed an adaptive QA framework, dynamically selecting optimal RAG strategy from simple to sophisticated through estimating query complexity by a trained smaller LLM.
\citet{zhang2024raft} proposed a retrieval-augmented fine-tuning strategy, enabling LLM to identify distractor documents and adapt to domains.
As for datasets, there are several multi-hop QA corpora, such as HotpotQA \cite{yang-etal-2018-hotpotqa} and WikiHop \cite{yang-etal-2018-hotpotqa}, which are widely used in RAG tasks.

\begin{figure*}
    \centering
    \includegraphics[width=0.95\linewidth]{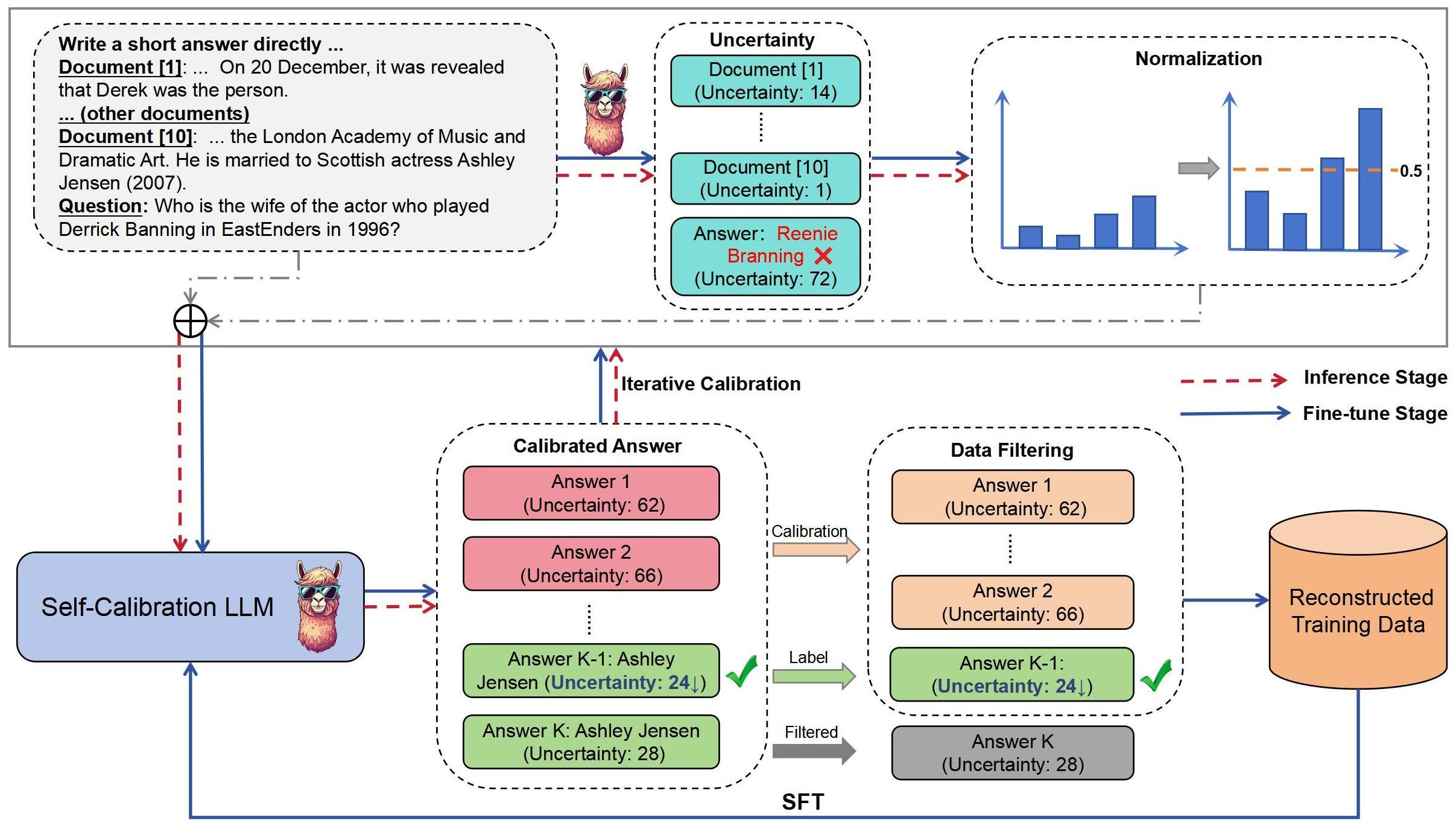}
    \caption{The overview of our framework. }
    \label{fig:2}
\end{figure*}

\subsection{Self-Calibration}
Self-calibration refers to LLMs learning from automated feedback to improve their behavior and adapt over time, avoiding costly human feedback \cite{pan2023automatically}. Traditional approaches \cite{li2019deep,unanue2021berttune, wu2021textgail} utilize meticulously designed external matrix to measure the generative generation quality of LMs to guide the model to perform the self-calibration. \citet{akyurek2023rl4f, yan2023learning, li2024think} further expanded these methods by modifying the matrix or introducing external revising modules. 
To avoid the increasingly complex matrix design, researchers construct frameworks such as CoT to calibrate the LLMs leveraging their strong inference ability iteratively\cite{zelikman2022star,wang2022selfi,wang2022selfc}. In addition, several methods \cite{du2023improving,li2024can,liang2024debatrix} employ multi-LLM debating frameworks to calibrate model-generated answers. However, these approaches require fine-tuning multiple LLM agents or utilizing LLMs with extremely large parameters, such as ChatGPT, for many rounds of interaction. In contrast, our approach requires fine-tuning the LLMs only once, enabling them to perform comprehensive reasoning and self-calibration based on retrieved documents, previously generated answers, and their associated uncertainty scores.

\section{Method}
Figure \ref{fig:2} provides an overview of our proposed framework, which comprises three main components: (1) estimating the uncertainty scores of each document and the generated answers (Section \ref{sec.2.1}); (2) iteratively utilizing the generated answers and their corresponding uncertainty scores from the validation set to perform the self-calibration process during the inference stage (Section \ref{sec.2.2}); and (3) designing a strategy to reconstruct a new training set to fine-tune a self-guided iterative calibration LLM with uncertainty awareness (Section \ref{sec.2.3}).

\subsection{Uncertainty Estimation} \label{sec.2.1}
To achieve our self-guided iterative calibration framework, we first employ the uncertainty score of the generated answer to evaluate the necessity of calibration according to the observation in Figure \ref{fig:1}. For a given input sequence $X$=[$x_1$,$x_2$, ...,$x_n$] consisting of $n$ tokens, the large language model generates a corresponding output sequence $Y$=[$y_1$,$y_2$, ...,$y_m$] with $m$ tokens, accompanied by the corresponding token-level logits. We begin by applying the softmax function to the logits to derive the highest probability associated with each token, denoted as $P$=[$p_1$, $p_2$, \ldots, $p_m$]. The uncertainty is then estimated by computing the product of these maximum probabilities, a widely adopted method for uncertainty estimation:
\begin{equation} \label{eq1}
    s_{ans} = (p_1 \times p_2 \times ... \times p_m)
\end{equation}

To ensure the consistency of this score across both the training and inference stages, we redefine Equation \ref{eq1} as follows:

\begin{equation} \label{eq2}
\begin{aligned}
   & \hat{s} =
    \begin{cases}
     \displaystyle\frac{s_{ans} - \overline{s}_{ans}}{1 - \overline{s}_{ans}} & \text{if } s_{ans} < \overline{s}_{ans} \\
     \displaystyle\frac{s_{ans} - \overline{s}_{ans}}{1 - \overline{s}_{ans}} & \text{otherwise}
    \end{cases} \\
   &s'_{ans} = 100 \times \left( 0.5 + 0.5 \times \hat{s} \right)
\end{aligned}
\end{equation}

where $\overline{s}_{ans}$ is the average uncertainty scores calculated by all generated answers in the training, validation, or test set seperately.

Furthermore, we incorporate uncertainty scores to evaluate the model’s confidence regarding the relevance of each document to the given question. A lower uncertainty score indicates a higher likelihood of the document being relevant, thereby aiding the LLM in information retrieval. Inspired by \citet{duan2023shifting}, we estimate the uncertainty score of each document by calculating the product of the maximum probabilities of the generated answers, obtained by combining the given question with each document individually.

\begin{equation} \label{eq3}
    \begin{split}
    s_{doc} &= 1-(p_1 \times p_2 \times ... \times p_m) \\
    s'_{doc} &= 100 \times \frac{(s_{doc}^i - \text{Min}(s_{doc}))}{(\text{Max}(s_{doc}) - \text{Min}(s_{doc}))}
    \end{split}
\end{equation}
where $s_{doc}^i$ is the product of maximum probability with the $i$-th document subtracted from one. A higher product indicates the LLM is more confident that the document is relevant to the question and contains more adequate information. To estimate the uncertainty score, we subtracted this product from one and normalized the uncertainty scores of all documents to reduce sensitivities.

Notably, the uncertainty estimation method can be replaced with any improved metric that more accurately measures uncertainty values. We also discuss the precision of uncertainty estimation for SGIC in Appendix \ref{app.7}.

\begin{table}[t]
    \small
  \centering
    \renewcommand\arraystretch{1}
    \scalebox{1}{
    \begin{tabular}{l}
    \toprule
    \textbf{Write a short answer ...} \\
    \textbf{Document [1]}: Yellowcraig, less commonly ... \\
    \textbf{... (other documents)} \\
    \textbf{Document [10]}: Dirleton Castle ... \\
    \hdashline
    \textbf{Question}: A medieval fortress in Dirleton,   ... \\
    \textbf{Previous Generated Answer}: \\
    Round 1: Lord High ... (Uncertainty Score: 73) \\ 
    Round 2: United States ...(Uncertainty Score: 51)\\
    \textbf{Answer}:  \\
      \bottomrule
    \end{tabular}}
    \caption{The input format of iterative self-calibration in our framework.}
  \label{table10}
\end{table}

\subsection{Iterative Self-Calibration} \label{sec.2.2}
During the inference stage, we primarily generate an answer based on the full documents and ascertain its uncertainty score employing Equation \ref{eq2}. Subsequently, we appraise the uncertainty score associated with the answer generated by each document following Equation \ref{eq3}. 
Then, we rephrase the input with these elements. As demonstrated in Figure \ref{fig:2}, we integrate the documents with corresponding uncertainty scores, while attaching the primary answer combined with uncertainty scores as the first-round original answer. The reformulated input is supplied to the LLM, which then engages in in-context reasoning and calibrates the answer under the directive of the uncertainty scores.


Furthermore, self-calibration is iteratively conducted $K$ rounds to obtain a satisfactory answer with minimal uncertainty. Specifically, in each round, the generated answer is re-input into the model for further calibration. During each calibration, we calculate the uncertainty scores of the answer according to Equation \ref{eq2} and incorporate the answer and score into the ``Previous Generated Answer'' as illustrated in Table \ref{table10} to reconstruct the input of subsequent round, enabling the answer to be calibrated iteratively based on the given documents and the previously generated answers.

\begin{table*}[t]
  \centering
    \renewcommand\arraystretch{0.95}
    \tabcolsep=0.8cm
    \scalebox{0.9}{
    \begin{tabular}{lcccc}
    \toprule
     & \multicolumn{2}{c}{\textbf{HotpotQA}} & \multicolumn{2}{c}{\textbf{Natural Question (NQ)}} \\
    \midrule
     Model& EM  & F1 & EM & F1 \\
     \midrule
     \multicolumn{5}{c}{Close-Source LLMs} \\
     \midrule
     GPT-4o-mini & 69.2 & 63.0 & 62.9 & 47.5 \\
     GPT-4o-mini (Ours) & \textbf{74.1}  & \textbf{66.7} & \textbf{64.4} & \textbf{48.8} \\
     \midrule
     GPT-4o & 73.7 & 68.1 & 63.3 &  53.0 \\
     GPT-4o (Ours) & \textbf{76.5} & \textbf{70.8} &  \textbf{65.2} & \textbf{55.0} \\
   
      \midrule
       \multicolumn{5}{c}{Open-Weight LLMs} \\
       \midrule
      Phi-3.5-mini (Full Tuning) & 42.8 & 50.3 & 58.7 & 64.3 \\
      Phi-3.5-mini (Ours) & \textbf{55.3} & \textbf{60.2} & \textbf{65.0} &  \textbf{67.5} \\
      \midrule
      Llama2-7B-Chat (LoRA Tuning) & 69.1 & 73.5 & 74.7 & 77.9 \\
      Llama2-7B-Chat (Ours) & \textbf{77.2} & \textbf{80.5} & \textbf{79.0} & \textbf{81.2} \\      
      \bottomrule
    \end{tabular}}
    \caption{The main experimental results (\%) on the dev set of HotpotQA and Natural Question (NQ) datasets. \textbf{Bold} numbers indicate the better result for each baseline, which sampling $K$ times following the iterative calibration process.}
  \label{table1}
\end{table*}

\subsection{Uncertainty-aware Fine-tuning} \label{sec.2.3}

After quantifying the uncertainty scores for the training data, we reformulate the representation of each data sample to incorporate these uncertainty scores, as illustrated in Figure \ref{fig:2}. Using this reformulation, we construct an uncertainty-aware self-calibration dataset derived from the original training corpus. Following the pipeline outlined in Section \ref{sec.2.2}, we restructure the input by including the content and uncertainty scores of documents alongside the primary answers.

The LLM is then employed to iteratively calibrate the answers for each data sample until the answer is correct or the round limit $k$ is reached. After $k$ rounds of calibration, samples with incorrect answers are removed. For the remaining samples, the final reformulated input is used as the training set input. Since the final input contains multi-round information in the ``Previous Generated Answer'' field, it provides the LLM with comprehensive knowledge of the calibration process. This pruning of unsuitable samples allows the LLM to focus on learning how to leverage the additional information encoded in the uncertainty scores for improved calibration. In parallel, a substitution operation is carefully designed to refine the model’s ability to address answers with high uncertainty while preserving or optimizing answers with low uncertainty. This prevents the model from erroneously altering correct answers.

The refined training set enhances the model’s ability to calibrate responses accurately while better capturing relevant information through the uncertainty scores. After the training set is restructured, we apply a standard supervised fine-tuning (SFT) procedure to fine-tune the self-guided iterative calibration LLM and evaluate its performance as described in Section \ref{sec.2.2}.

\begin{table}[t]
    \centering
    \renewcommand\arraystretch{1}
    \scalebox{0.9}{
    \begin{tabular}{cccc}
        \toprule
        Dataset & Train & Validation & Test \\
        \midrule
        HotpotQA & 50,000 & 7,405 & 7,405  \\
        NQ & 40,000 & 2,000 & 2,000 \\ 
        \bottomrule
    \end{tabular}}
    \caption{The statistics of \textbf{HotpotQA} and \textbf{Natural Question (NQ)} in our experimental setting.}
    \label{tab:1}
\end{table}

\section{Experiments}
\subsection{Setup}
\paragraph{Dataset} We evaluate our proposed framework on \textbf{HotpotQA} \cite{yang-etal-2018-hotpotqa} and \textbf{Natural Question} (NQ) \cite{kwiatkowski-etal-2019-natural} corpus. The statistic of these two datasets is shown in Table \ref{tab:1}. \textbf{HotpotQA}\footnote{https://hotpotqa.github.io/} dataset includes 113k multi-hop questions. There are two types of questions: bridge and comparison. The final answer in the distractor setting is generated through 10 passages. Each question has at least 2 relevant passages. We also reconstruct the \textbf{Natural Question} (NQ)\footnote{https://github.com/google-research-datasets/natural-questions} \cite{kwiatkowski-etal-2019-natural} dataset in the distractor setting similar to \textbf{HotpotQA} to evaluate the robustness of our framework. The question of the NQ dataset is comprised of a Google query and a corresponding Wikipedia page. Each page has a passage that can answer the question. We take nine other passages from the same page as distractors. Because of the high demand for computational resources in LLM and the computing resource limitations, we use part of the training set in our experiments.

\begin{table}[t]
  \centering
    \renewcommand\arraystretch{1}
    \tabcolsep=0.54cm
    \scalebox{0.9}{
    \begin{tabular}{lcc}
    \toprule
     Question Type & EM  & F1  \\
     \midrule
    Bridge & 65.0 & 72.9 \\
      Bridge (Ours) & \textbf{75.7} & \textbf{79.4} \\
      \midrule
      Comparison & 69.6 & 73.2 \\
      Comparison (Ours) & \textbf{83.1} & \textbf{84.8} \\
      \bottomrule
    \end{tabular}}
    \caption{The results of different types of question (\%) in HotpotQA dataset on Llama2-7B-Chat. \textbf{Bold} numbers indicate the best results.}
  \label{table7}
\end{table}

\paragraph{Evaluation} For all experiments in this work, we employ two widely used metrics, Exact Match (EM) and F1 score, to evaluate the performance of our framework. We perform calibration $K$ times on samples that require calibration. From these iterations, we select the response with the minimal uncertainty score as the final correct answer.

\paragraph{Models} For closed-source LLMs, we evaluate our proposed method on the \textbf{GPT-4o-mini} and \textbf{GPT-4o} models \cite{hurst2024gpt} through the OpenAI API without fine-tuning on the downstream datasets. As for open-weight baselines, we employ two strong decoder-only large language models (LLMs) that vary in scale and architecture: \textbf{Phi-3.5} \cite{textbooks2} and \textbf{Llama2-7B-Chat} \cite{touvron2023llama}. The implementation of experiments will be explained in Appendix \ref{appendix.2}.

\subsection{Main Results}
As shown in Table \ref{table1}, we show our main experimental results on two closed-source LLMs and two open-weight LLMs, which are evaluated with HotpotQA and Natural Question (NQ) datasets. For close-source LLMs, which are GPT-4o-mini and GPT-4o, our method consistently achieves better performance on both EM and F1 scores compared with sampling the response $K$ times directly. As for the open-weight LLMs, the Phi-3.5-mini and Llama2-7B-Chat models, which are fine-tuned with our proposed framework, also outperform the baselines on all EM and F1 scores, whether with the LoRA tuning or full tuning. 

To explore the effectiveness of our approach on different types of problems, we also comparatively analyze the effectiveness of our approach on bridge and comparison types of questions within the dev set of HotpotQA dataset, which are shown in Table \ref{table7}. Our framework obtained more than 10\% improvements on both two types of questions. This suggests that our approach works not only on comparison questions with binary answers, but also on bridge questions with open answers, which demonstrate the robustness and generality of our method. 

\begin{table}[t]
  \centering
    \renewcommand\arraystretch{0.9}
    \tabcolsep=0.2cm
    \scalebox{0.9}{
    \begin{tabular}{lcc}
    \toprule
     Model& EM  & F1  \\
    \midrule
      Llama2-7B-Chat (LoRA Tuning) & 69.1 & 73.5 \\
      \hdashline
      \textbf{+} Calibration & 71.8 & 75.3 \\
        \textbf{+} Calibration  \& Answer \\ \quad Uncertainty & 76.2 & 79.6 \\
        \textbf{+} Calibration  \& Document \\ \quad Uncertainty (Ours) & \textbf{77.2} & \textbf{80.5} \\
      \bottomrule
    \end{tabular}}  
    \caption{The ablation study results (\%) on the dev set of HotpotQA dataset. \textbf{Bold} numbers indicate the best results. }
  \label{table3}
\end{table}

\subsection{Ablation Study}
To evaluate the various components of our proposed framework, we conducted a series of experiments to assess the performance impact of combining: (a) only the initial answer, (b) the initial answer with its uncertainty score, and (c) the uncertainty scores of documents. It should be noted that, for experiment (a), we selected the calibrated answer repeated in the second iteration as the final answer, following \citet{huang2024large}, since uncertainty scores are not available when only the initial answer is used.
As shown in Table \ref{table3}, the combination of most components yields superior results in both EM and F1 evaluation metrics, demonstrating the effectiveness of our proposed methodology.

\section{Analysis}
\begin{table}[t]
  \centering
    \renewcommand\arraystretch{0.9}
    \tabcolsep=0.23cm
    \scalebox{0.9}{
    \begin{tabular}{lcc}
    \toprule
     Model& EM  & F1  \\
    \midrule
      Llama2-7B-Chat (LoRA Tuning) & 69.1 & 73.5 \\
      External Relevant Score & 75.4 & 78.8 \\
     \chen{Oracle Uncertainty}  & \textbf{85.7} & \textbf{85.1} \\   
      \hdashline
      Llama2-7B-Chat (Ours) & 77.2 & 80.5 \\
      \bottomrule
    \end{tabular}}
    \caption{The experimental results (\%) on the dev set of HotpotQA dataset with different settings of the uncertainty scores in our method. \textbf{Bold} numbers indicate the best results.}
  \label{table4}
\end{table}

\subsection{The Impact of Uncertainty Scores}
Despite the depiction of the observed uncertainty distribution in Figure \ref{fig:1}, we conducted experiments to validate the necessity of this pattern within our framework. As evidenced in Table \ref{table4}, we replaced the uncertainty scores of the documents with the relevant scores calculated by the \textbf{Multi-qa-mpnet-base-dot-v1} model, a powerful document extraction model proposed by \citet{reimers-2019-sentence-bert}. We believe this is because the uncertainty scores computed by the model itself are strongly correlated with the model's capabilities, and therefore, more effectively guide the model in iterative self-calibration. Moreover, we consider an extreme case where the uncertainty scores for relevant/irrelevant documents and correct/wrong answers are perfectly accurate. In this scenario, we assign uncertainty scores ranging from 0 to 20 for relevant documents and correct initial answers, while irrelevant documents and wrong answers were given uncertainty scores spanning from 80 to 100. The EM scores improved to 85.7, and the F1 scores improved to 85.1, which further justifies the potential of our framework. Furthermore, we provide a theoretical analysis regarding the application of uncertainty scores in Appendix \ref{app.6}.

\begin{figure}[t]
    \centering
    \includegraphics[width=1\linewidth]{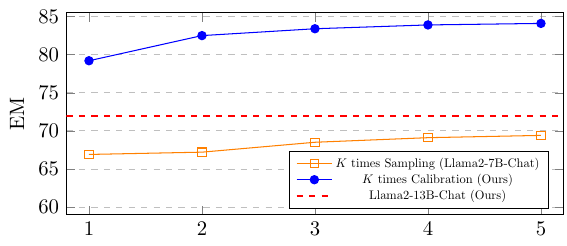}
    \caption{The EM scores of the Llama2-7B-Chat fine-tuned with our method to perform the $K$ times calibrations on the dev set of the HotpotQA dataset.}
    \label{fig:3}
\end{figure}

\subsection{The Performance of Iterative Calibration}
We conducted additional experiments to thoroughly investigate the implications of the $K$-times calibration mechanism. The blue solid curve in Figure \ref{fig:3} illustrates the experimental outcomes with $K=\{1,2,3,4,5\}$ iterations of self-calibration. Intuitively, the performance improves as the number of calibration iterations increases. Remarkably, it even surpasses the fine-tuned Llama2-13B-Chat model and the same model fine-tuned with our framework, demonstrating the potential of our approach. For a fair comparison, we also performed 1 to 5 sampling iterations with a temperature of 0.7 for the fine-tuned Llama2-7B-Chat model, represented by the solid orange line in Figure 3. As expected, the performance of the model gradually improves with an increasing number of samples. However, it is evident that our framework still shows a significant performance advantage over this baseline. Additionally, we compare the inference cost of our self-calibration method with the sampling baseline in Appendix \ref{app.4}.

\subsection{Uncertainty Change of Iterative Calibration}
\begin{figure}[t]
    \centering
    \includegraphics[width=1\linewidth]{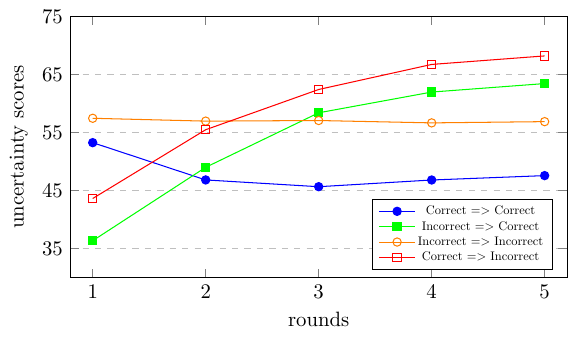}
    \caption{The uncertainty scores of each round of calibrations in four modes of answer variation.}
    \label{fig:7}
\end{figure}

To clarify the iterative calibration process, we present the average uncertainty scores for how answers change in each calibration round, as illustrated in Figure \ref{fig:7}. The uncertainty for ``Correct => Correct'' and ``Incorrect => Incorrect'' remains relatively stable, which makes sense since the answers do not change in these cases. However, the uncertainty scores for ``Correct => Incorrect'' and ``Incorrect => Correct'' gradually increase with more calibrations. This likely happens because as more ``Previous Generated Answers'' accumulate, the model engages in more complex reasoning, leading to higher uncertainty. Interestingly, the uncertainty for ``Incorrect => Correct'' samples is consistently lower than for ``Correct => Incorrect'' samples in each round, indicating that the model exhibits lower uncertainty when the calibration results in a correct answer. This can help the model recognize the successful calibrations.

\begin{figure}[t]
    \centering
    \includegraphics[width=1\linewidth]{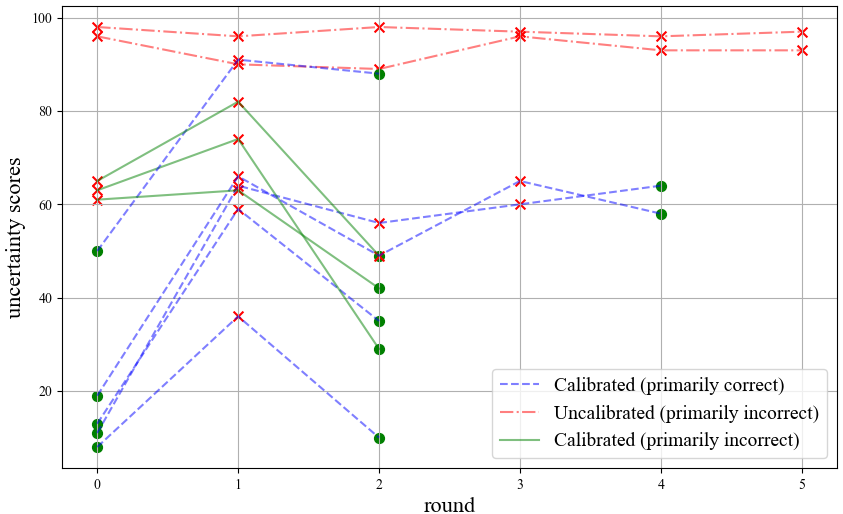}
    \caption{The answer correctness and uncertainty scores of 10 random samples during the iterative calibration. Round 0 refers to the initially generated answer without document uncertainty. {\color{red}{Red}} samples refers to incorrect answers, while {\color{green}{Green}} samples refers to correct answers.}
    \label{fig:6}
\end{figure}

In more detail, we also conduct a more concrete analysis of the 10 randomly extracted samples and results in each round to visualize changes in their answer correctness and corresponding uncertainty scores throughout our calibration procedure. The result is demonstrated in Figure \ref{fig:6}, which indicates three types of calibration procedures. For the samples that remain uncorrected after five times calibrations, the uncertainty scores persist at high levels during the whole process. This reveals the LLM is confused by the question and documents. In contrast, for those samples calibrated within five times, the LLM starts with being suspicious of the incorrect answer, as indicated by an increase in uncertainty scores in the first time calibration. The LLM then adjusts the suspected answer in the following iteration with a lower uncertainty score. Moreover, we observe that some initially correct answers are altered to be incorrect in the first round, As evidenced by a significant rise in uncertainty scores, the alternations resemble uncertain attempts. Subsequently, the LLM identifies the errors and calibrates the answers in subsequent iterations.

\begin{table}[t]
  \centering
    \renewcommand\arraystretch{0.9}
    \tabcolsep=0.54cm
    \scalebox{0.9}{
    \begin{tabular}{lc}
    \toprule
     Model & EM   \\
     \midrule
    Llama2-7B-Chat (Fine-tuned) & 36.2 \\
      Llama2-7B-Chat (Ours) & \textbf{40.1} \\
      \bottomrule
    \end{tabular}}
    \caption{The experimental results (\%) on the test set of GSM8K dataset. \textbf{Bold} numbers are the best results for each base model.}
  \label{table8}
\end{table}

\subsection{Generalizability}
In addition to evaluating our approach to RAG tasks, we conduct experiments on other types of reasoning tasks to demonstrate their broader applicability. Specifically, we employ the \textbf{GSM8K} dataset \cite{cobbe2021training}, which contains basic mathematical problems necessitating multi-step reasoning, to fine-tune both the baseline and our method on the Llama2-7B-Chat model. The GSM8K dataset comprises 7,473 training instances and 1,319 testing instances. For evaluation, we compute the Exact Match (EM) score using the final computation result, without considering the intermediate reasoning steps. Additionally, we treat each step in the reasoning process generated by the pre-trained LLM as a document in the RAG task to calculate the uncertainty score, which is then used for fine-tuning and self-calibration.

As presented in Table \ref{table8}, fine-tuning the model by reconstructing the input to match the format shown in Table \ref{table9} yields an approximate 4\% improvement in model performance with only one-time calibration. This consistent enhancement highlights the generalizability of our approach and its potential applicability to a range of reasoning tasks.

\begin{table}[t]
  \centering
    \renewcommand\arraystretch{0.9}
    \tabcolsep=0.38cm
    \scalebox{0.9}{
    \begin{tabular}{lcc}
    \toprule
     Dataset & EM  & F1  \\
     \midrule
     \multicolumn{3}{c}{Fine-tuned on NQ} \\
     \midrule
    HotpotQA (Sampling 5 times) & 54.5 & 57.5 \\
      HotpotQA (Ours) & \textbf{71.0} & \textbf{74.4} \\
      \midrule
     \multicolumn{3}{c}{Fine-tuned on HotpotQA} \\
     \midrule
      NQ (Sampling 5 times) & 58.7 & 67.4 \\
      NQ (Ours) & \textbf{76.5} & \textbf{78.8} \\
      \bottomrule
    \end{tabular}}
    \caption{The results of Llama2-7B-Chat on the test set of one dataset, fine-tuned on the other. \textbf{Bold} numbers indicate the best results.}
  \label{table11}
\end{table}

\subsection{Transferability}
We investigate the transferability of calibration capabilities in LLMs fine-tuned with our framework. Specifically, we assess the impact of using LLMs fine-tuned on one dataset when applied to the dev set of another dataset. As shown in Table \ref{table11}, sampling five times with the LLM fine-tuned on a different dataset results in significantly lower performance compared to the baseline fine-tuned on the corresponding training set (Table \ref{table1}). This occurs because, in the original QA task, LLMs primarily learn to extract answers from documents, often overlooking in-context reasoning abilities. Additionally, the single-hop of NQ and the multi-hop of HotpotQA influence the LLMs' transferability. However, our framework's self-calibration capability outperforms the baseline and nearly matches the performance of LLMs fine-tuned with the specific training set. This is due to our approach's emphasis on reasoning through multiple rounds of answers and documents, using uncertainty scores to identify more plausible answers, which is a common logic applicable across all datasets.

\subsection{Error Analysis}
\chen{
To explore deeper into the distinctions between our method and baseline, we performed an error analysis using the HotpotQA dataset. In the baseline, error samples with low uncertainty are typically more likely to be calibrated successfully, whereas those with high uncertainty often fail calibration. Our analysis of calibration performance across 7,405 samples showed that 1,440 samples (19.5\%) were successfully calibrated, while 1,080 samples (14.6\%) failed. Among the successful calibrations, 75\% (1,081 samples) were calibrated within the first two rounds. This was due to the high initial uncertainty in the baseline answers, which prompted divergent responses. The remaining 359 samples required iterative calibration: baseline answers with low uncertainty gradually accumulated doubt over rounds until a threshold was reached, prompting revised responses. Correct answers reinforced confidence, reducing uncertainty, while incorrect ones sustained high uncertainty, prompting further revisions. For the failures, 693 cases retained incorrect answers with declining uncertainty. A small subset proposed alternative incorrect answers accompanied by sharp drops in uncertainty, halting calibration. Notably, 387 failures reached the 5-round limit with rising uncertainty, suggesting potential calibration success if the rounds were extended. This indicates the framework’s ability to iterative self-calibration, although round limits currently constrain efficacy.}

\subsection{Case Study}

We present two examples in Table \ref{table5} to empirically showcase the capabilities of SGIC. In Example 1, our uncertainty-aware self-calibration for LLMs calibrates errors in generated answers using uncertainty scores from source documents and the initial answer. In Example 2, we illustrate the framework’s ability to refine partially correct answers, enhancing both the completeness and accuracy of the responses. These examples provide compelling evidence of the effectiveness of our framework in improving the quality and reliability of LLM-generated answers.

\begin{figure}[t]
    \centering
    \includegraphics[width=0.9\linewidth]{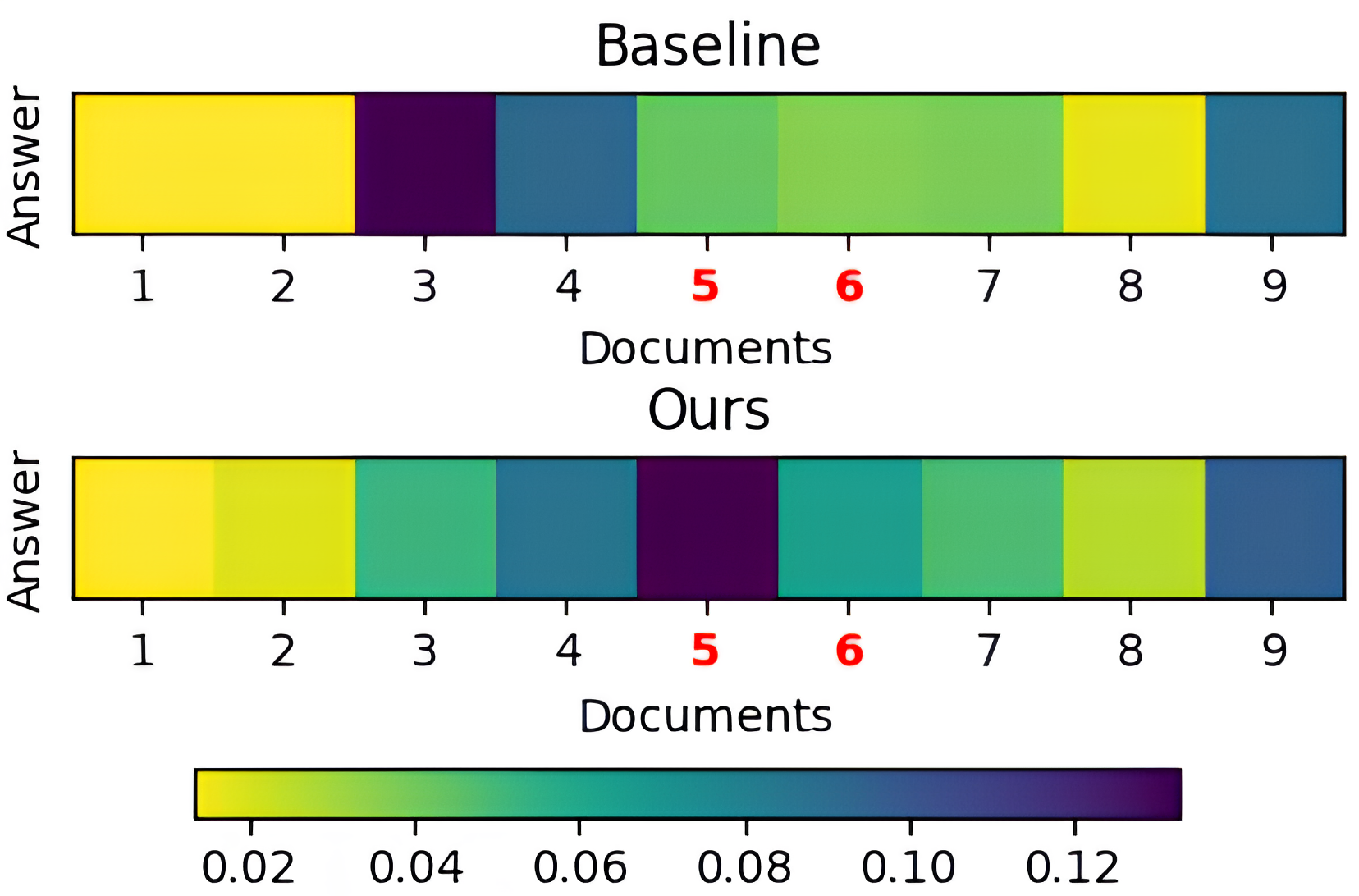}
    \caption{The visualization of the attention distribution of the given documents in a sample. {\color{red}{Red}} means the relevant two documents.}
    \label{fig:4}
\end{figure}

\subsection{Attention Distribution}
To further investigate how our framework enables LLMs to self-calibrate, we visualize the attention distribution across various documents to analyze its impact on LLM parameterization. Figure \ref{fig:4} presents a comparison between the attention allocation of Llama2-7B-Chat fine-tuned with our framework (lower) and the baseline model (upper). Notably, the baseline model exhibits low attention distribution to relevant documents, whereas our self-calibration framework directs attention more effectively toward these relevant documents. This observation indicates that our framework enhances the model's ability to calibrate its answers by focusing attention on pertinent information.

For a more detailed evaluation, we examined the attention distribution for the whole test set of the HotpotQA dataset. Since the relevant documents may appear in different positions among all candidates, we quantified the difference using the $R_{10}@k$ score. This metric represents the proportion of the top two relevant documents ranked within the top k positions, based on the attention distribution in the LLM. We compared the attention weights from the baseline model and after the first calibration. The results, displayed in Table \ref{table13}, reveal that even a single calibration with our framework significantly improves the model's focus on relevant documents compared to the baseline.

\section{Conclusion}
In this paper, we observed significant gaps in uncertainty scores from various LLMs and estimation methods between relevant/irrelevant documents and correct/incorrect answers in the RAG task. To address this, we proposed a novel framework that guides iterative calibration using the model's in-context reasoning abilities. Our framework consistently improves performance for both open-weight and closed-source models by utilizing uncertainty scores of documents and generated answers. These findings underscore the potential of uncertainty-aware self-calibration in enhancing the accuracy and reliability of large language models.

\section{Limitation}
Even though our proposed novel SGIC framework can utilize the in-context reasoning capabilities of large language models (LLMs) to iteratively self-calibrate the answers by leveraging the uncertainty scores of the given documents and the initially generated answers, it is contingent upon the precision of the underlying uncertainty estimation. While much current work has devoted considerable attention to methodologies for gauging the uncertainty indicative of the veracity of generated responses, there remains a dearth of studies addressing the assessment of the confidence regarding the pertinence of documents to the question. In the future, we will further explore how to improve the accuracy of uncertainty estimation, which can maximize the effectiveness of our framework.

\section*{Acknowledgments}
This work was supported in part by the Science and Technology Development Fund of Macau SAR (Grant Nos. FDCT/0007/2024/AKP, FDCT/0070/2022/AMJ, FDCT/060/2022/AFJ), the National Natural Science Foundation of China (Grant Nos. 62261160648, 62266013), the China Strategic Scientific and Technological Innovation Cooperation Project (Grant No. 2022YFE0204900), and the UM and UMDF (Grant Nos. MYRG-GRG2023-00006-FST-UMDF, MYRG-GRG2024-00165-FST-UMDF, EF2024-00185-FST, EF2023-00151-FST, EF2023-00090-FST).

\bibliography{acl_latex}

\appendix

\section{Appendix}
\label{sec:appendix}

\subsection{Implementation Details} \label{appendix.2}
We use the LLaMA-Factory\footnote{https://github.com/hiyouga/LLaMA-Factory} GitHub repository to fine-tune the \textbf{Llama2-7B-Chat} with LoRA \cite{hu2021lora, DBLP:journals/corr/abs-2503-23360} and \textbf{Phi-3.5-mini} with full-parameters mode because of the limitation of computation resource. We train all models using the batch size of 16. We set the initial learning rate to 5e-5 and fine-tuned all models 3 epochs. We truncate each document and ensure that its length is less than 200 tokens for all the open-weight LLMs. All the experiments have been completed on one 80G H800 GPU or 80G A100 GPU. Besides, all experiments are calibrated at most five rounds. During the calibration phase, we ensure an equitable evaluation by sampling $k$ times for the baseline model the same as an equivalent number of experimental trials with the baseline model.

\begin{figure}[t]
    \centering
    \includegraphics[width=1\linewidth]{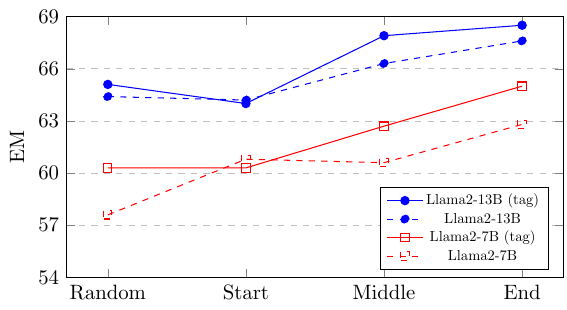}
    \caption{The experimental results of using the ``$<$KEY$>$'' tag to guide LLM on the RAG task at different order of documents. The solid line is the result of adding the ``$<$KEY$>$'' tag, while the dashed line is the baseline result.}
    \label{fig:5}
\end{figure}

\subsection{The Impact of the Tag In Context} \label{sec.tag}
 Firstly, we leverage large language models (LLMs) without fine-tuning to conduct experiments aimed at evaluating the effectiveness of using specific tags as hints within the context to guide LLM reasoning. Table \ref{table6} illustrates the input format utilized in these experimental settings. We selected a random subset of 2,000 instances from the development set of the HotpotQA dataset. Each instance adds the ``$<$key$>$'' tags before the relevant documents.

To mitigate potential biases arising from input sequencing in LLMs, as identified by \citet{liu2023lost}, we systematically varied the positions of the given documents. In separate trials, the two relevant documents are placed at the beginning, middle, and end of the input sequence, as well as in random positions. The results of these experiments are depicted in Figure \ref{fig:5}. Notably, the use of tags to direct the model's attention to relevant documents demonstrates an improvement in performance on retrieval-augmented generation (RAG) tasks. This finding supports our hypothesis that the model has the potential to calibrate its generated answers through in-context reasoning guided by specific tags.

\begin{table}[t]
  \centering
    \renewcommand\arraystretch{1}
    \tabcolsep=0.3cm
    \scalebox{1}{
    \begin{tabular}{lcccc}
    \toprule
     Dataset & \multicolumn{2}{c}{HotpotQA} & \multicolumn{2}{c}{NQ} \\
        \midrule
      & EM  & F1 & EM & F1  \\
     \midrule
      Self-RAG & 48.7 & 29.9 & 61.3 & 70.8 \\
      Ours & \textbf{77.2} & \textbf{80.5} & \textbf{79.0} & \textbf{81.2} \\
      \bottomrule
    \end{tabular}}
    \caption{The results of comparing our method with the Self-RAG framework.}
  \label{table12}
\end{table}

\subsection{Analysis of Inference Cost} \label{app.4}
\chen{
One of the concerns regarding the self-calibration framework is its inference cost, as it requires generating an initial answer and then calibrating it. To address this, we conducted experiments to analyze the inference cost of our proposed method. Initially, we compared the first calibration cost with the baseline implementation of Llama2-7B-Chat fine-tuned with LoRA, using the same number of sampling iterations. Our framework retains 72\% of the baseline's inference speed while achieving superior performance, as demonstrated in Figure \ref{fig:3}. This empirically validated trade-off highlights our framework's ability to deliver significant performance improvements with minimal latency increase.

Additionally, we conducted a comparative analysis with one of the state-of-the-art frameworks, Self-RAG \cite{asai2023self}. The experimental results, detailed in Table \ref{table12}, provide significant insights. Using the Llama2-7B model as per Self-RAG's setup, our framework, with 10 retrieved documents, shows notable improvements over Self-RAG. Specifically, our SGIC framework enhances Exact Match (EM) scores by 28.5 on the HotpotQA dataset and 17.7 on the Natural Questions (NQ) dataset.

A key strength of our framework is its enhanced computational efficiency, achieved through batch processing of document uncertainties. On the HotpotQA dataset, SGIC achieves 210\% of Self-RAG's inference speed. Although this speed advantage decreases to 90\% on the NQ dataset, this reduction is due to Self-RAG's shorter generation pattern on the simpler single-hop NQ dataset, which reduces time spent on retrieval, critique, and generation phases. Furthermore, Self-RAG requires full parameter fine-tuning and relies on GPT-4 to generate extensive critique training data, increasing training overhead. In contrast, our framework eliminates the need for external knowledge and accelerates training using Parameter-Efficient Fine-Tuning (PEFT) methods like LoRA, achieving superior results compared to Self-RAG.
}


\subsection{Theoretical Analysis} \label{app.6}
\chen{In this section, we delve into the theoretical underpinnings of why our method is effective. The Probability Ranking Principle (PRP), as introduced by \citet{robertson1977probability}, posits that ranking documents based on their probability of relevance ensures optimal performance in ad-hoc retrieval tasks. This principle relies on models that provide well-calibrated probability estimates. Recent studies have shown that integrating uncertainty into ranking processes can significantly boost learning and performance in information retrieval scenarios \cite{yu2024dynamic}. Our method adheres to the PRP by enhancing model calibration through robust uncertainty estimation.

While \citet{duan2023shifting} points out challenges such as generative inequalities in uncertainty quantification, our research emphasizes how uncertainty can be harnessed to improve retrieval models. By fine-tuning models to effectively utilize uncertainty scores, we achieve outcomes that align with the PRP, thereby demonstrating the reliability of our approach. Furthermore, employing advanced estimation techniques has the potential to further elevate performance, highlighting the robustness and versatility of our method. Beyond retrieval, our approach is applicable to long document challenges, such as translation and summarization, showcasing its adaptability across a wide range of tasks.}

\begin{table}[t]
  \centering
    \renewcommand\arraystretch{1}
    \tabcolsep=0.38cm
    \scalebox{1}{
    \begin{tabular}{lcc}
    \toprule
     Method &  $R_{10}@2$ & $R_{10}@5$  \\
     \midrule
        Baseline & 42.9 & 70.1 \\
      1st Time Calibration & \textbf{49.8} & \textbf{75.4} \\
      \bottomrule
    \end{tabular}}
    \caption{The $R_{10}@k$ scores of ranking the candidate documents from largest to smallest in terms of the weight of the attention distribution in Llama2-7B-Chat on the HotpotQA dataset.}
  \label{table13}
\end{table}

\subsection{Discussion of Uncertainty Estimation} \label{app.7}
\chen{In this work, our primary focus is on leveraging uncertainty scores to identify the best-calibrated answers by comparing these scores, rather than solely concentrating on the precision of uncertainty estimation. Nonetheless, we also assess the precision of uncertainty estimation for both our method and the baseline using a consistent estimation framework. We utilize the AUROC score as a metric and evaluate it on the HotpotQA test set. Following the 1st time calibration, our approach achieves an AUROC score of 68.4, surpassing the baseline's score of 65.5. This indicates that our method not only excels in selecting well-calibrated answers but also enhances uncertainty estimation, even though this is not our primary objective.}

\begin{table*}[t]
  \centering
    \renewcommand\arraystretch{0.9}
    \scalebox{0.9}{
    \begin{tabular}{l}
    \toprule
    \textbf{Write a short answer directly ...} \\
    \textbf{$<$KEY$>$ means this document contains key information of the question.} \\

    \textbf{(Other docuemtns) ...} \\
    \textbf{Document [6]}: {\color{blue}{$<$KEY$>$}} Yellowcraig, less commonly known as Broad Sands Bay, is a coastal area ... \\
    \textbf{Document [7]}: {\color{blue}{$<$KEY$>$}} Dirleton Castle is a medieval fortress in the village of Dirleton, East ... \\
      \bottomrule
    \end{tabular}}
    \caption{The input format of adding {\color{blue}{$<$KEY$>$}} tags.}
  \label{table6}
\end{table*}

\begin{table*}[t]
  \centering
    \renewcommand\arraystretch{1}
    \scalebox{0.9}{
    \begin{tabular}{l}
    \toprule
    \textbf{Explain your reasoning and give a final answer for the given question. The initial} \\
    \textbf{Question}: Josh decides to try flipping a house.  He buys a house for \$80,000 and ...\\ 
    \textbf{Reasoning Steps}: \\
    1. The house was worth \$80,000 before the repairs.(Uncertainty Score: 57)\\
    2. The repairs cost \$50,000.(Uncertainty Score: 53) \\
    3. So the house is now worth \$80,000 + \$50,000 = \$<<80000+ ...
    (Uncertainty Score: 4)\\
    4. The repairs increased the value of the house by 150\%.(Uncertainty Score: 36) \\
    5. So the value of the house increased by 150\% of \$50,000 = \$<<150 ... (Uncertainty Score: 100) \\
    6. So the house is now worth \$130,000 - \$75,000 = \$<<130000-75000= ... (Uncertainty Score: 0) \\
   7. So Josh made a profit of \$55,000. (Uncertainty Score: 49) \\
    \textbf{Final Answer}:  \\
    
      \bottomrule
    \end{tabular}}
    \caption{The input format of fine-tuning our method on the GSM8K dataset. }
  \label{table9}
\end{table*}

\begin{table*}[t]
  \centering
    \renewcommand\arraystretch{1}
    \scalebox{0.9}{
    \begin{tabular}{l}
    \toprule
    \multicolumn{1}{c}{Example 1} \\
    \midrule
    \textbf{Write a short answer directly without any explanation or introduction for the} ... \\ 
    \textbf{(Documents and their uncertainty scores)} ... \\
    \textbf{Question}: A medieval fortress in Dirleton, East Lothian, Scotland borders on the south side of ...\\
    \textbf{The initial answer is}: \\ 
     (Previous Rounds Generated Answer) \\
    Round 2: {\color{red}{Firth of Forth }}(Uncertainty Score: 90) \\
    \textbf{The correct answer is}: {\color{green}{Yellowcraig}} \\
    \midrule
    \multicolumn{1}{c}{Example 2} \\
    \midrule
    \textbf{Write a short answer directly without any explanation or introduction for the}... \\ 
     \textbf{(Documents and their uncertainty scores) ...} \\
    \textbf{Question}: The director of the romantic comedy ``Big Stone Gap'' is based in what New York city? \\
    \textbf{The initial answer is}: \\
    (Previous Rounds Generated Answer) \\
    Round 3: {\color{blue}{Greenwich Village }}(Uncertainty Score: 66) \\
    \textbf{The correct answer is}: {\color{green}{Greenwich Village, New York City}}\\     
      \bottomrule
    \end{tabular}}
    \caption{Two examples of how our proposed framework corrects the answer. {\color{red}{Red}} means the answer is wrong, while {\color{green}{green}} indicates the answer is correct. {\color{blue}{Blue}} indicates that the answer contains partially correct answer.}
  \label{table5}
\end{table*}

\end{document}